%% file: paper.tex
\pdfoutput=1
\documentclass[letterpaper,twocolumn,fleqn]{article}

\usepackage{ist}
\usepackage{amsmath}
\usepackage{amssymb}
\usepackage{subfig}
\usepackage{float}
\usepackage{color}

\DeclareMathAlphabet{\mathcal}{OMS}{cmsy}{m}{n} 

\def\I{\mathbf{I}}
\def\M{\mathbf{M}}
\def\Mh{\hat{\mathbf{M}}}
\def\Mt{\tilde{\mathbf{M}}}

\def\P{\mathbf{P}}

\title{Satellite Image Forgery Detection and Localization Using GAN and One-Class Classifier}

\author{Sri Kalyan Yarlagadda; School of Electrical and Computer Engineering, Purdue University; West Lafayette, IN, USA\\
	David G\"{u}era; School of Electrical and Computer Engineering, Purdue University; West Lafayette, IN, USA\\
	Paolo Bestagini; Dipartimento di Elettronica, Informazione e Bioingegneria, Politecnico di Milano; Milano, Italy\\
	Fengqing Maggie Zhu; School of Electrical and Computer Engineering, Purdue University; West Lafayette, IN, USA\\
	Stefano Tubaro; Dipartimento di Elettronica, Informazione e Bioingegneria, Politecnico di Milano; Milano, Italy\\
	Edward J. Delp; School of Electrical and Computer Engineering, Purdue University; West Lafayette, IN, USA}

\date{} 

\begin{document}

\maketitle

\thispagestyle{empty} 


\begin{abstract}
Current satellite imaging technology enables shooting high-resolution pictures of the ground.
As any other kind of digital images, overhead pictures can also be easily forged.
However, common image forensic techniques are often developed for consumer camera images, which strongly differ in their nature from satellite ones (e.g., compression schemes, post-processing, sensors, etc.).
Therefore, many accurate state-of-the-art forensic algorithms are bound to fail if blindly applied to overhead image analysis.
Development of novel forensic tools for satellite images is paramount to assess their authenticity and integrity.
In this paper, we propose an algorithm for satellite image forgery detection and localization. Specifically, we consider the scenario in which pixels within a region of a satellite image are replaced to add or remove an object from the scene.
Our algorithm works under the assumption that no forged images are available for training.
Using a generative adversarial network (GAN), we learn a feature representation of pristine satellite images.
A one-class support vector machine (SVM) is trained on these features to determine their distribution.
Finally, image forgeries are detected as anomalies.
The proposed algorithm is validated against different kinds of satellite images containing forgeries of different size and shape.
\end{abstract}

\input{intro}
\input{rel_work}
\input{method}
\input{results}
\input{conclusions}

\section{Acknowledgments}
This material is based on research sponsored by DARPA and Air Force Research Laboratory (AFRL) under agreement number FA8750-16-2-0173. The U.S. Government is authorized to reproduce and distribute reprints for Governmental purposes notwithstanding any copyright notation thereon. The views and conclusions contained herein are those of the authors and should not be interpreted as necessarily representing the official policies or endorsements, either expressed or implied, of DARPA and Air Force Research Laboratory (AFRL) or the U.S. Government.



\small
\bibliographystyle{IEEEbib}
\bibliography{paper.bbl}


\begin{biography}
Sri Kalyan Yarlagadda received his btech in Electrical Engineering from Indian Institute of Technology Madras in July 2015. He is currently pursuing a Ph.D. at the School of Electrical and Computer Engineering, Purdue University, USA. His research interests are video/image processing and deep learning. \\

David G\"{u}era received his B.E. (Hons.) in Telecommunications Systems Engineering from Polytechnic University of Catalonia - BarcelonaTech, Spain in July 2016. He is currently pursuing a Ph.D. at the Video and Image Processing Laboratory (VIPER), School of Electrical and Computer Engineering, Purdue University, USA. His research interests are video/image processing and deep learning. \\

Paolo Bestagini received the M.Sc. in Telecommunications Engineering and the Ph.D. in Information Technology at the Politecnico di Milano, Italy, in 2010 and 2014, respectively. Since 2016 he is Assistant Professor at the Department of Electronics, Informatics and Bioengineering, Politecnico di Milano. His research activity is focused on multimedia forensics and acoustic signal processing for microphone arrays. \\

Fengqing Zhu is an Assistant Professor of Electrical and
Computer Engineering at Purdue University, West Lafayette, IN.
Dr. Zhu received her Ph.D. in Electrical and Computer Engineering
from Purdue University in 2011. Prior to joining Purdue
in 2015, she was a Staff Researcher at Huawei Technologies
(USA), where she received a Huawei Certification of Recognition
for Core Technology Contribution in 2012. Her research interests
include image processing and analysis, video compression,
computer vision and computational photography.\\

Stefano Tubaro completed his studies in Electronic Engineering at the Politecnico di Milano, Italy, in 1982. He joined the Politecnico di Milano, first as a researcher and in 1991 as an Associate Professor. Since 2004 he has been appointed as Full Professor of Telecommunications. His current research interests are on advanced algorithms for video and sound processing and for image and video tampering detection. He authored over 180 scientific publications and more than 15 patents. \\

Edward J. Delp was born in Cincinnati, Ohio. He is currently The Charles William Harrison Distinguished Professor of Electrical and Computer Engineering and Professor of Biomedical Engineering at Purdue University. His research interests include image and video processing, image analysis, computer vision, image and video compression, multimedia security, medical imaging, multimedia systems, communication and information theory. Dr. Delp is a Life Fellow of the IEEE, a Fellow of the SPIE, a Fellow of IS\&T, and a Fellow of the American Institute of Medical and Biological Engineering.\\

\end{biography}

\end{document}

%% file: intro.tex
\section{Introduction}\label{sec:intro}

Ever since the birth of the Internet, the accessibility to images has become easier overtime.
Internet has become an affordable and effective platform for distributing one's own images.
User friendly software like Photoshop and Gimp can be used to generate a variety of image manipulations such as inpainting, copy-forge, splicing, etc.
A combination of the above two scenarios is a perfect environment for producing doctored images, which when treacherously used can cause substantial damage.
Therefore, it is of paramount importance to develop forensic methods to validate the integrity of an image.
For this reason, over the years, the forensic community has developed several techniques for image authenticity detection and integrity assessment \cite{Rocha2011, Piva2013, Stamm2013}.

In addition to photographs captured with cameras and smartphones, other types of imagery are starting to be circulated, posing new problems for the forensic community.
Indeed, current satellite imaging technology enables shooting high-resolution pictures of the ground.
Due to the increased availability of satellites equipped with imaging sensors, overhead images are becoming popular.
It is now possible to easily gather overhead images of the ground through public websites \cite{freesat} and to buy custom image sets of specific locations and times.
As any other kind of digital images, overhead pictures can also be easily forged.
One question that needs to be addressed is whether these images are authentic.
Cases of malicious overhead image manipulations have already been reported \cite{bbc}, \cite{russia}. The development of forensic methods tailored to the analysis of this type of imagery is considered to be urgent.

However, common image forensic techniques are often developed for consumer cameras, which strongly differ in their nature from satellite sensors (e.g., compression schemes, post-processing, sensors, etc.).
Therefore, many accurate state-of-the-art forensic algorithms are bound to fail if blindly applied to overhead image analysis.
Development of novel forensic tools for satellite images is paramount to assess their authenticity and integrity.

To fill the lack of ad-hoc forensic techniques for satellite images, the authors of \cite{Ho2005} proposed an active method based on watermark embedding.
Watermarks can then be exploited to detect possible doctored image regions.
Unfortunately, this method can only be used if watermark is inserted at image inception time.
More recently, the authors of \cite{Ali2017} proposed a passive forensic method for overhead image analysis.
This algorithm is based on machine learning techniques, but it can only localize image regions that have been inpainted.
To the best of our knowledge, no specific algorithms for other kinds of satellite image forgeries have been proposed in the literature.

In this paper, we propose an algorithm for satellite image forgery detection and localization.
Specifically, we consider the situation in which pixels within a region of a satellite image are replaced to add or remove an object from the scene.
Our algorithm works under the assumption that no forged images are available for training.
Using a generative adversarial network (GAN), we learn a feature representation of pristine satellite images.
A one-class support vector machine (SVM) is trained on these features to determine their distribution. Finally, image forgeries are detected as anomalies.

To validate the proposed method, we built a custom dataset of forged satellite images using different forgery sizes.
Results in terms of forgery detection and localization are presented. Moreover, as the proposed algorithm works by analyzing images patch-wise, it is possible to strongly parallelize it to keep processing time at bay.

%% file: rel_work.tex
\section{Problem Definition and Background}
In this section we describe the problem formulation and notation used throughout the entire paper.
Following this, we provide some background concepts on autoencoders and convolutional neural networks.

\subsection{Problem formulation}
\begin{figure}[t]
	\centering
	\subfloat[Pristine image]{\includegraphics[width=.4\columnwidth]{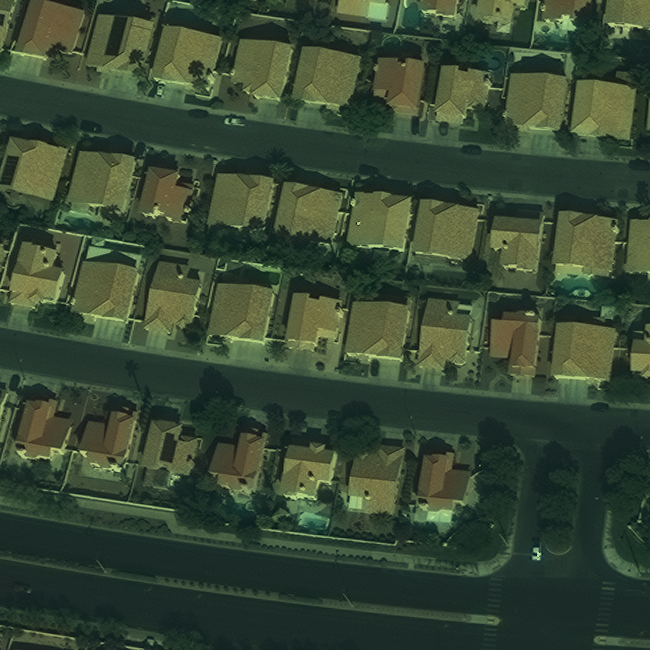}}\hfil
	\subfloat[Pristine mask]{\includegraphics[width=.4\columnwidth]{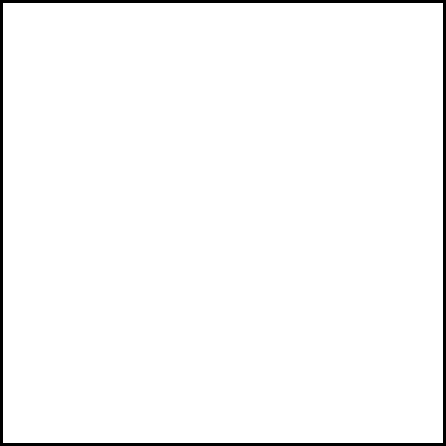}}

	\subfloat[Forged mask]{\includegraphics[width=.4\columnwidth]{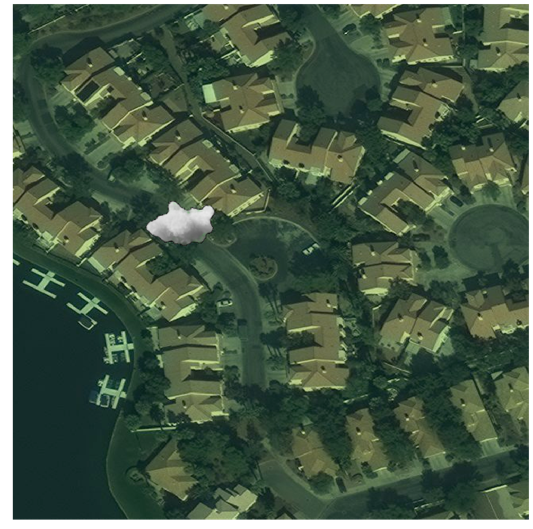}}\hfil
	\subfloat[Forged mask]{\includegraphics[width=.4\columnwidth]{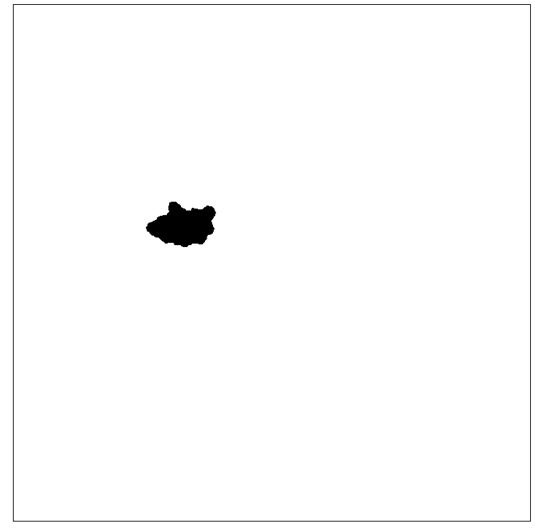}}
	\vspace{1em}
	\caption{Example of pristine (a) and forged (c) images $\I$ associated to their binary forgery masks $\M$ (b) and (d), respectively}
	\label{fig:problem}
\end{figure}

Consider an image \textbf{I} coming from a satellite.
We can represent the pixel integrity associated with the image $\I$, as a binary mask $\M$ of the same size as the image in pixels.
Each entry of $\M$ is a binary label $0$ or $1$, such that a pixel belonging to a forged area is assigned the label $0$ and a pixel from an untampered area is assigned a $1$. As forgery, in this paper we consider an object insertion / removal through a copy-paste operation from a different source. This means that forged pixels do not belong to a satellite image but come from a different device (e.g., the picture of a plane acquired with a normal camera). Figure~\ref{fig:problem} shows an example of a pristine satellite image and a completely white (i.e., label $1$) mask, as well as a forged image with the respective black and white mask localizing the forgery.
Within this setup, our goal is twofold:
\begin{itemize}
	\item \textit{Tampering Detection}: given an image, detect whether it is pristine or forged.
	\item \textit{Tampering Localization}: given a forged image, detect which are the forged pixels.
\end{itemize}
These two tasks can be accomplished by computing $\Mh$ (i.e., an estimate of $\M$). If $\Mh$ contains any entry different from $1$, the image is detected as forged. Entries of $\Mh$ whose values are $0$ represent forged pixel positions.

%

\subsection{Related Work}
In this section we present a brief summary of autoencoders that are needed to follow this paper. For a thorough review, we recommend the readers to refer to Chapter 14 of \cite{Goodfellow2016}.

Autoencoders are neural networks that are trained to attempt to obtain an output equal to the input through a set of linear and non-linear operations that expand or reduce data dimensionality at some point in the network. They consist of two parts: the encoder $A_e$ and decoder $A_d$. The output of the encoder is called feature vector or hidden representation, and we represent it as $\textbf{h}$. In this paper we work with autoencoders where the dimensionality of $\textbf{h}$ is lower than the dimensionality of the input. This kind of autoencoders are known as undercomplete autoencoders. From now on, whenever we refer to autoencoders we refer to undercomplete autoencoders. Such an architecture forces the autoencoder to capture a salient representation of the input in a reduced dimensionality space.

Autoencoders are trained by minimizing through iterative procedures a loss value defined as
\begin{equation}
	L = \mathcal{L}(x,\; A_d(A_e(x))),
\end{equation}
where $\mathcal{L}(\cdot,\; \cdot)$ is the loss function computing some distance between its two arguments, $x$ is the autoencoder input, and $A_d(A_e(x))$ is the output. In the special case where $\mathcal{L}$ is the mean squared error (MSE) loss, the autoencoder learns to perform a generalized non linear principal component analysis (PCA).

In this paper we design our autoencoders using Convolutional Neural Networks (CNNs).
CNNs have proven to be very successful in a variety of computer vision tasks such as object recognition \cite{DBLP:journals/corr/HeZRS15}, object detection \cite{DBLP:journals/corr/HeGDG17}, etc.
They came to limelight in 2012 \cite{NIPS2012_4824} when they produced stunning results in the ImageNet Large Scale Visual Recognition Challenge (ILSVRC) \cite{DBLP:journals/corr/RussakovskyDSKSMHKKBBF14}.
Since then there has been an explosion in the application of CNNs to various other computer vision tasks and often resulting in new state of the art results.

When it comes to image forensics, the use of CNNs has been on the rise.
Many forensic problems deal with non-linear and often difficult to model pipelines.
Therefore, CNNs have proven to be successful in this area.
The first works using CNNs in this area were focused on steganalysis \cite{qian2015deep, pibre2015deep, Xu2016, Sedighi2017}.
Strictly concerning multimedia forensics, many other tasks have been considered.
As a few examples, \cite{Chen2015} deals with median filtering detection, \cite{Bayar2016} proposes the use of a constrained convolutional layer for forgery detection.
In \cite{Tuama2016a, Bondi2017, Bondi2017a, Guera2017}, the problem of camera model identification and its possible forgeries is explored.
Double JPEG compression is also considered in \cite{wang2016double, Barni2017}.

Convolutional neural networks usually consist of operations such as convolutions, batch normalization \cite{BN}, local pooling, thresholding and non linear activations. These operations are stacked together and are tuned by minimizing a cost function at the output. Following, we describe some of the most commonly used layers:
\begin{itemize}
	\item
	Convolutional: the input of this layer is convolved with a bank of filters whose response is learned through training. The input is typically a 3D structure, i.e., it has two spatial coordinates plus depth (e.g., an RGB image). The output is known as feature map.
	\item
	Max pooling: given an input $x$, a sliding window is used to extract the maximum value over it.
	\item
	Batch Normalization: given an input $x$, this layer normalizes $x$ by imposing zero mean and unit variance. Details about this are explained in \cite{BN}.
	\item
	Deconvolutional: this layer is the transpose of a convolutional layer. The output is obtained by convolving a zero-padded version of the input with a filter bank learned through training.	The spatial dimensions of the output are greater than that of the input.
\end{itemize}

%% file: method.tex
\section{Method}
\begin{figure}[t]
	\centering
	\includegraphics[width=\columnwidth]{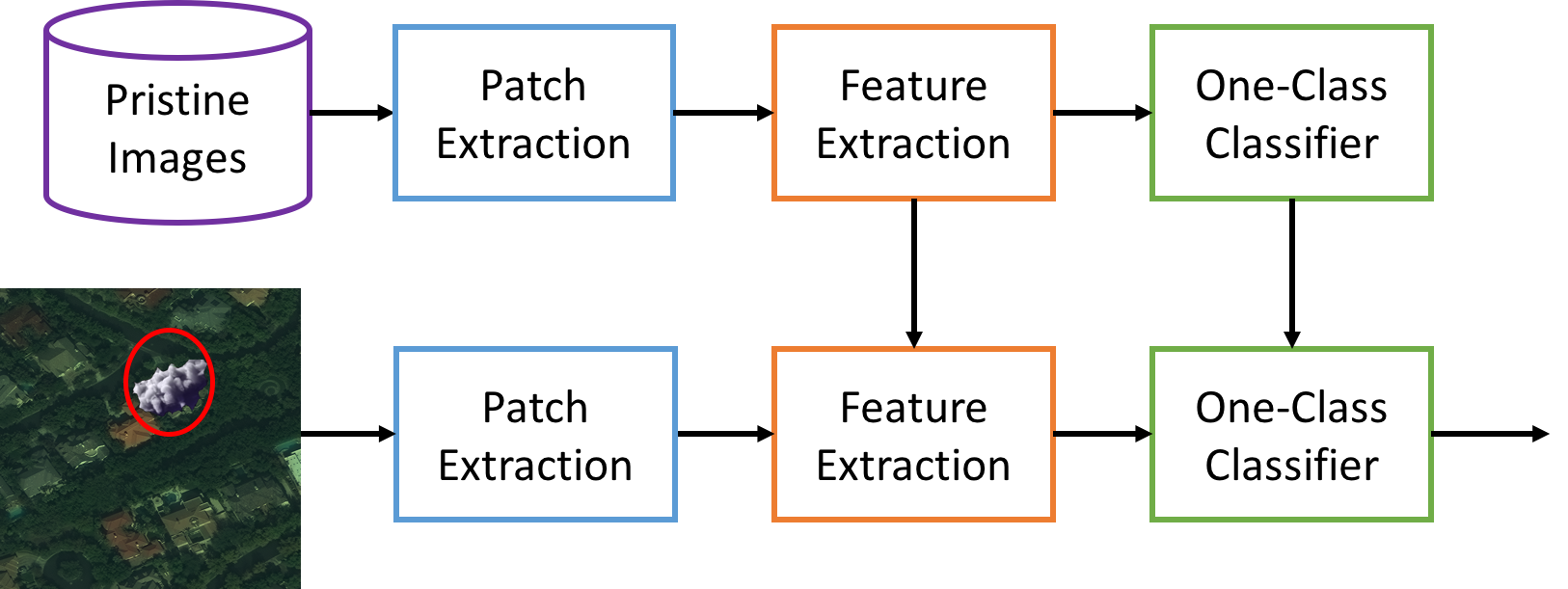}
	\vspace{1em}
	\caption{Pipeline of the proposed method. At training time, the feature extractor and one-class SVM learn their models from pristine images only. At testing time, forged areas are detected as anomaly with respect to the learned model. }
	\label{fig:pipeline}
\end{figure}

In this section we elaborate on the details of our method to detect object insertion / deletion attacks in satellite images. In particular, the pipeline of our method is reported in Figure~\ref{fig:pipeline}, and it is composed by the following steps:
\begin{itemize}
	\item The color image under analysis is split into patches (either overlapping or not) of size $64 \times 64$ pixels.
	\item A adversarially trained autoencoder encodes the patches into a low dimensional representation called feature vector $\textbf{h}$.
	\item A one-class SVM fed with $\textbf{h}$ is used to detect forged patches as anomalies with respect to features distribution learned from pristine patches.
	\item Once all patches are classified, a label mask for the entire image is obtained by grouping together all the patch labels.
\end{itemize}

The rationale behind the proposed solution is that autoencoders are able to capture a reduced dimensionality representation of the input data, still retaining important characteristic information, as shown in \cite{Cozzolino2016, DAvino2017} for forensic purposes. Therefore, by training an autoencoder only on pristine data, we expect it to learn to extract features specific of original satellite images. Conversely, when it is tested on forged data, the extracted features should be strongly different from those obtained from pristine images. A one-class SVM trained on pristine features only can then be used to discriminate between features coming from pristine and forged images. Following, we report a detailed explanation of each step of the proposed pipeline.

\subsection{Patch Extraction}
The given image $\I$ is split into regular patches $\P_{k}$, where $k \in [1, K]$ is the patch index, and $K$ is the total amount of patches. Patches can be either overlapped or not depending on the selected trade-off between detection accuracy and computational complexity (i.e., overlapping leads to more patches to analyze but more accurate results).

\begin{figure}[t]
	\centering
	\includegraphics[width=\columnwidth]{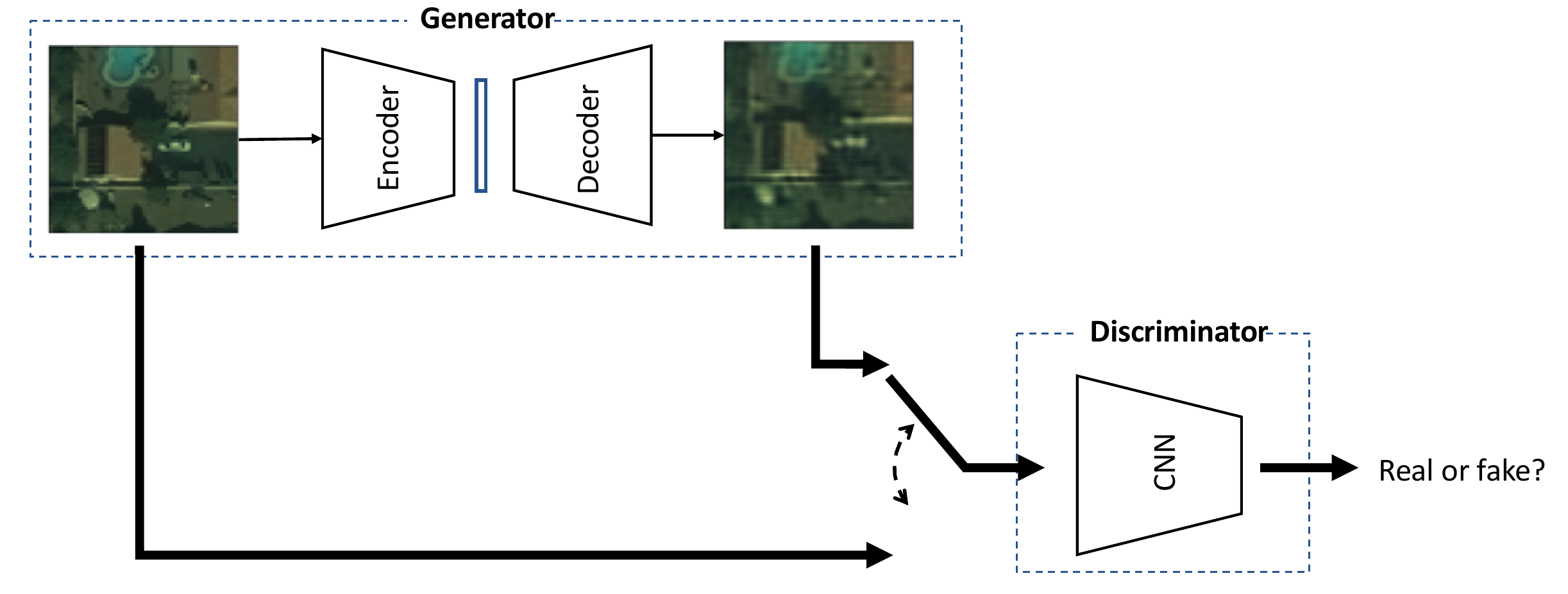}
	\vspace{1em}
	\caption{Architecture of the used GAN.}
	\label{fig:gan}
\end{figure}
\subsection{Feature Extraction}
Every patch $\P_k$ is fed to the autoencoder $A$ which consists of two parts: the encoder $A_e$ and decoder $A_d$. Both $A_e$ and $A_d$ are made of convolutional and deconvolutional neural networks respectively. They are symmetric in terms of the number of layers. The architecture of $A_e$ has been selected following the same rationale of \cite{Pathak2016} and it is as follows:
\begin{itemize}
	\item \textit{conv1}: convolution layer with 16 filters each of size (6,6) with stride 1.
	\item  \textit{conv2}: convolution layer with 16 filters each of size (5,5) with stride 2.
	\item  \textit{conv3}: convolution layer with 32 filters each of size (4,4) with stride 2.
	\item  \textit{conv4}: convolution layer with 64 filters each of size (3,3) with stride 2.
	\item  \textit{conv5}: convolution layer with 128 filters each of size (2,2) with stride 2.
\end{itemize}

All convolutional layers except \textit{conv5} are followed by batch normalization. All the convolution layers are activated using a linear function. The output of \textit{conv5} is the feature vector $\mathbf{h}$, a $2048$ dimensional vector and has a much lower dimension than that of the input which is  $12288$ dimensional. The architecture of $A_d$ is as follows:
\begin{itemize}
	\item \textit{dconv1}: deconvolution layer with 64 filters each of size (2,2) with stride 2.
	\item  \textit{dconv2}: deconvolution layer with 32 filters each of size (3,3) with stride 2.
	\item  \textit{dconv3}: deconvolution layer with 16 filters each of size (4,4) with stride 2.
	\item  \textit{dconv4}: deconvolution layer with 16 filters each of size (5,5) with stride 2.
	\item  \textit{dconv5}: deconvolution layer with 3 filters each of size (6,6) with stride 1.
\end{itemize}
Every deconvolutional layer is followed by batch normalization except \textit{deconv5}. \textit{Deconv5} has a hyperbolic tangent activation where all other deconvolution layers have linear activations. The output of \textit{deconv5} is the output of the autoencoder. Once the autoencoder is trained on pristine image patches, we use it as feature extractor to compute the feature vector $\mathbf{h}_k = A_e(\P_k)$ from each image patch $\P_k$.

Conventionally $A$ can be trained using stochastic gradient descent to minimize mean squared loss between input (i.e., $\P_k$) and output (i.e., $A_d(A_e(\P_k))$) . However better results can be achieved when we follow an adversarial framework for training the autoencoder. In \cite{NIPS2014_5423} the authors established a framework of min-max adversarial game between two neural networks, namely the generator and discriminator, and such networks are called Generative Adversarial Networks. As shown in Figure~\ref{fig:gan}, the discriminator aims to accurately discriminate between patches from real satellite images and patches created by the generator. The generator on the other hand aims to mislead the discriminator by trying to generate data  closer and closer to the real one. Such frameworks have proven to be extremely effective.

The architecture of the discriminator $D$ we use is as follows:
\begin{itemize}
	\item \textit{conv1}: convolution layer with 16 filters each of size (5,5) with stride 1 followed by Leaky ReLU and batch normalization.
	\item  \textit{conv2}: convolution layer with 16 filters each of size (2,2) with stride 2.
	\item  \textit{conv3}: convolution layer with 32 filters each of size (4,4) with stride 1 followed by Leaky ReLU and batch normalization.
	\item  \textit{conv4}: convolution layer with 32 filters each of size (2,2) with stride 2.
	\item  \textit{conv5}: convolution layer with 64 filters each of size (3,3) with stride 1 followed by Leaky ReLU and batch normalization.
	\item  \textit{conv6}: convolution layer with 64 filters each of size (2,2) with stride 2 followed by Leaky ReLU and batch normalization.
	\item  \textit{fc1}: A 128-neuron fully connected dense layer followed by a Leaky ReLU activation.
	\item \textit{fc2} : A single neuron followed by a sigmoid activation.
\end{itemize}

After training the autoencoder using the GAN strategy on pristine images only, the encoder $A_e$ is used to extract the feature vector $\textbf{h}_k$ from each patch $\P_k$ under analysis.

\subsection{One-Class SVM}
The autoencoder $A$ is trained only on pristine patches and hence it learns to encode them very well. So when $A$ sees a patch containing a forgery, it encodes it quite differently. In order to capture this difference without any knowledge on forged data, we use a one-class SVM trained on feature vectors $\mathbf{h}$ extracted from pristine images only. The used one class SVM learns pristine feature distribution. It then outputs a soft value which represents the likelihood of the feature vector $\mathbf{h}$ under analysis being pristine. We define the soft mask $\Mt$ as a matrix the same size of the image, where each entry contains the soft SVM output relative to the image patch in the same position. This soft mask $\Mt$ can be used to obtain the final detection binary mask $\Mh$ by simply thresholding.

%% file: results.tex
\section{Experimental Validation}

\begin{figure}[b!]
	\centering
	\subfloat[Image with small sized splicing]{\includegraphics[width=.4\columnwidth]{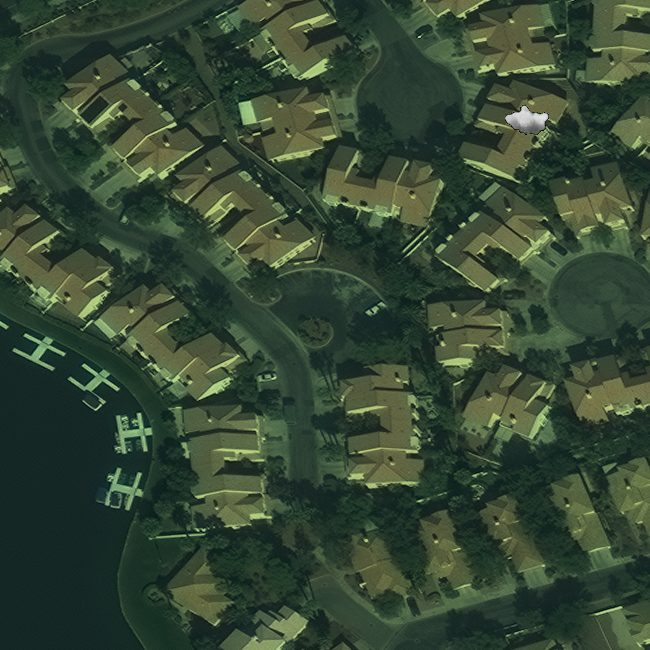}}\hfil
	\subfloat[Forged mask]{\includegraphics[width=.4\columnwidth]{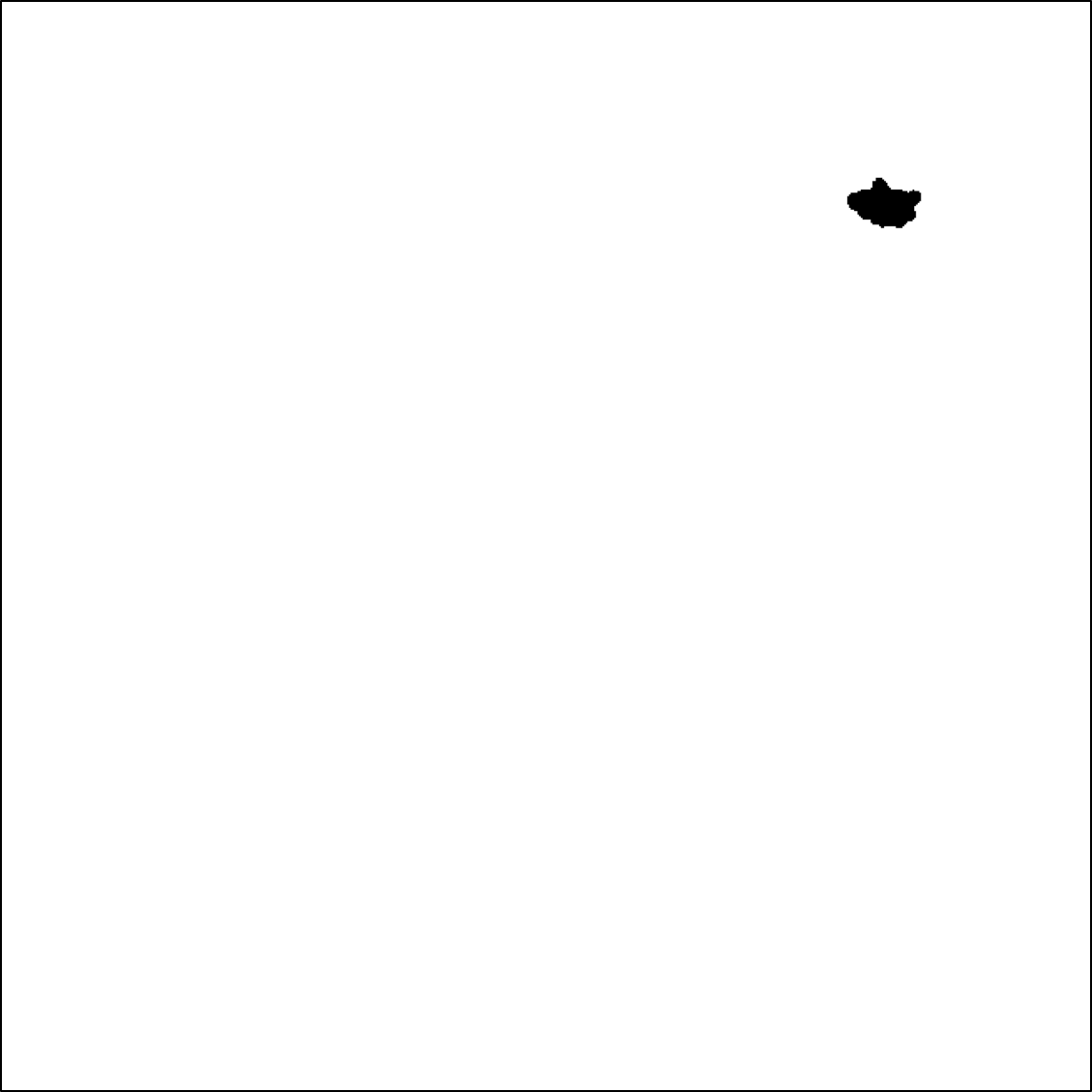}}

	\subfloat[Image with medium sized splicing]{\includegraphics[width=.4\columnwidth]{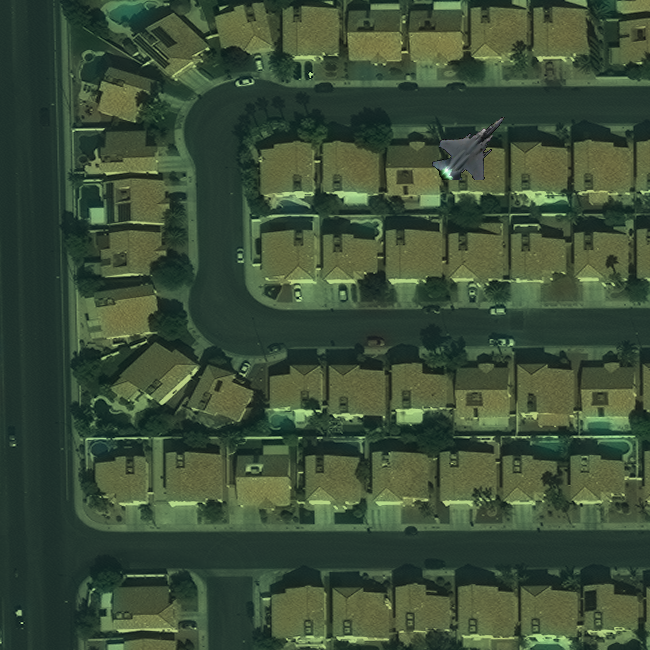}}\hfil
	\subfloat[Forged mask]{\includegraphics[width=.4\columnwidth]{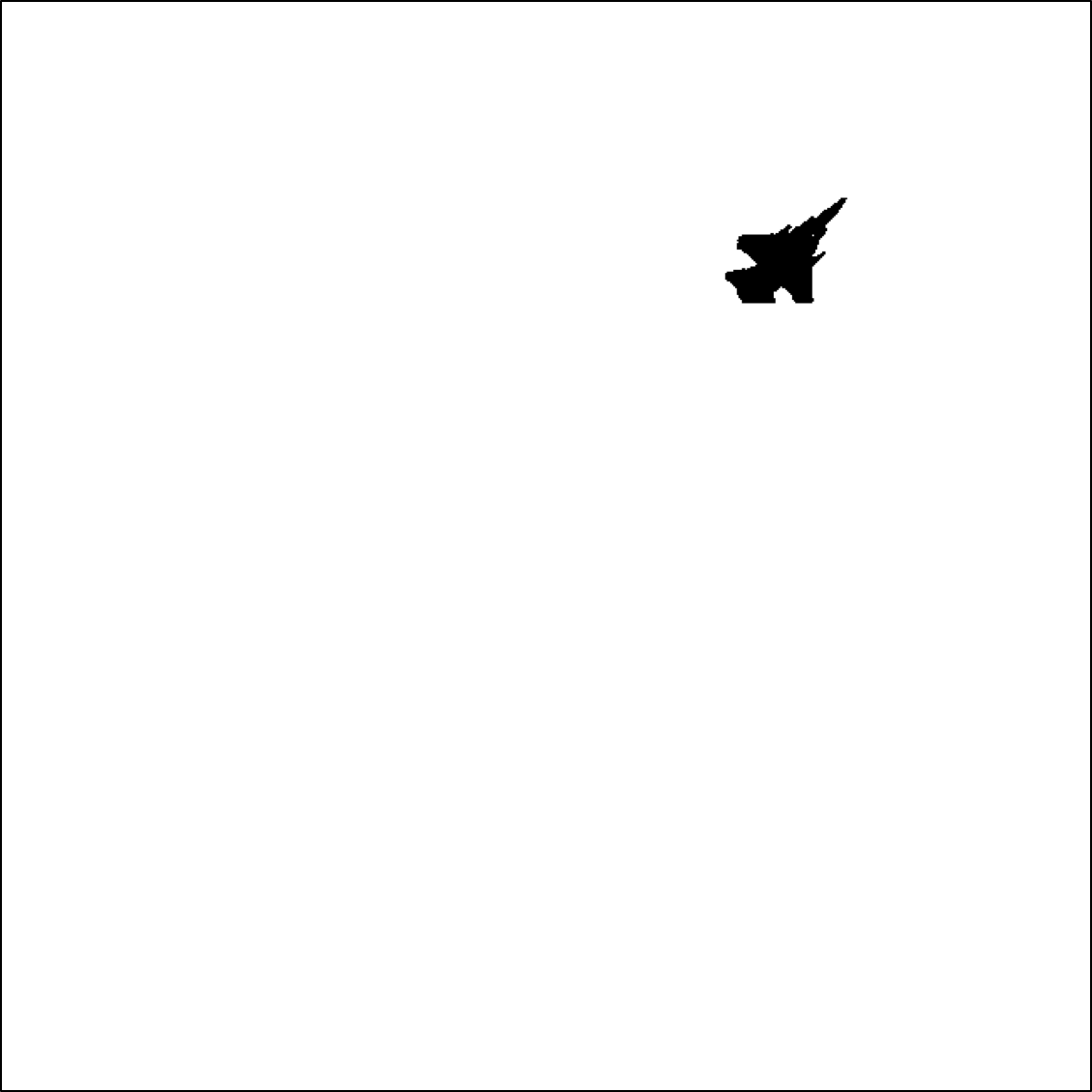}}

	\subfloat[Image with large sized splicing]{\includegraphics[width=.4\columnwidth]{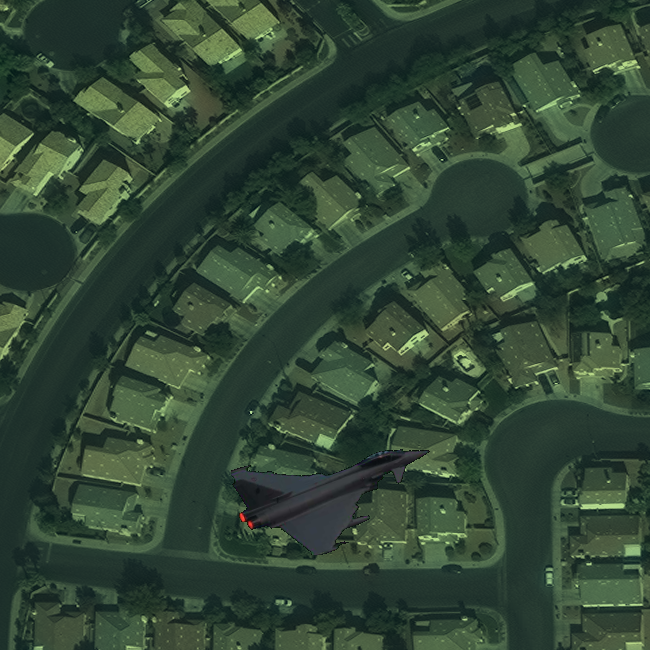}}\hfil
	\subfloat[Forged mask]{\includegraphics[width=.4\columnwidth]{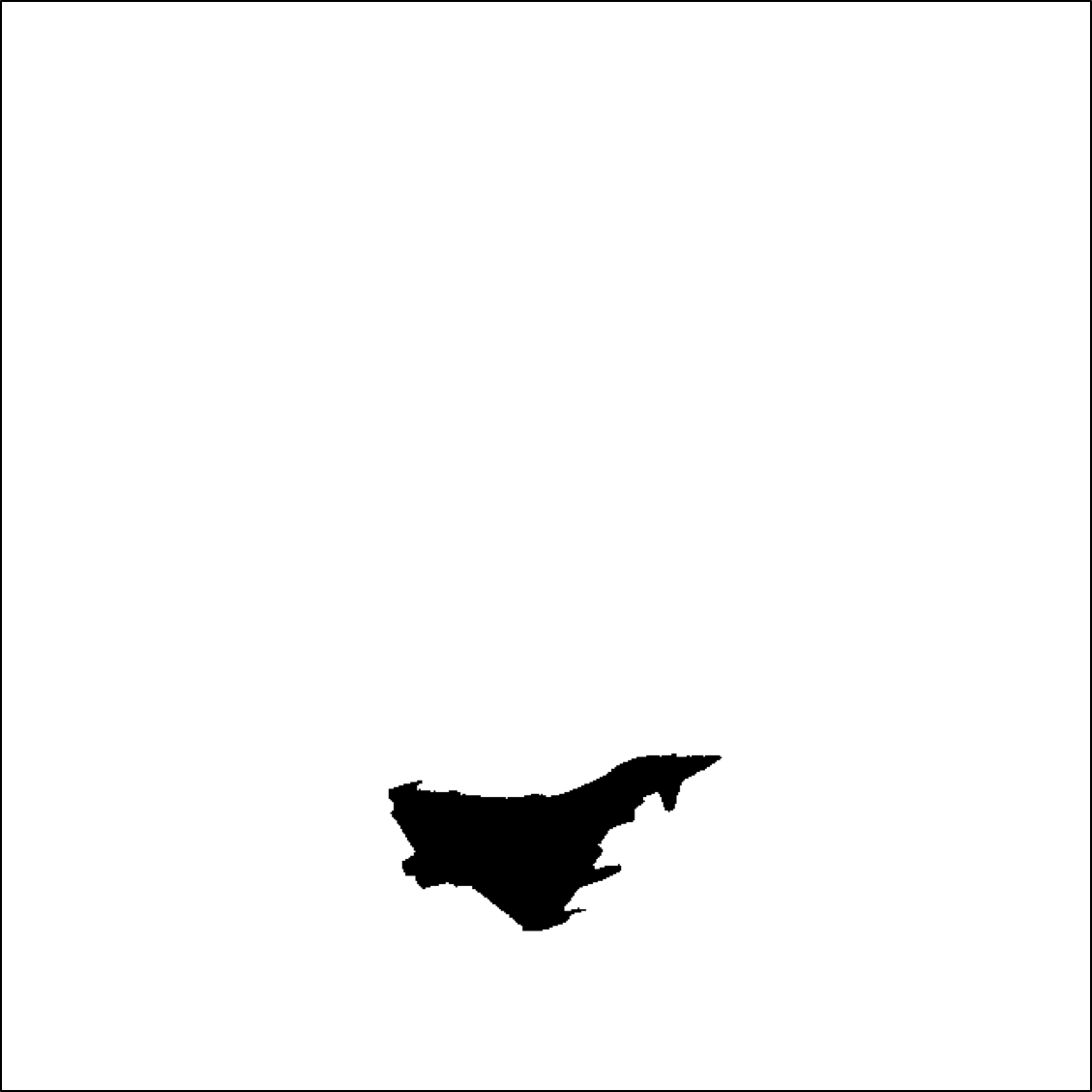}}

	\vspace{1em}
	\caption{Examples of forged images with forgeries of different sizes. Ground truth masks $\M$ are also reported.}
	\label{fig:dataset}
\end{figure}

In this section we report the experimental validation of the proposed technique. We first discuss how we built the used dataset. We then provide details about the considered experimental setup for reproducible research. Finally, we show the achieved numerical results.

\subsection{Dataset}

We tested our algorithm using overhead images obtained from the Landsat Science program \cite{landsat, landsat2}. The Landsat Science is a program run jointly by NASA \cite{NASA} and the US Geological Survey(USGS) \cite{USGS}. It was first launched in 1972 and has produced the longest, continuous record of Earth's land surface as seen from Space. NASA is responsible for the remote sensing equipment, launching satellites and validating their performance. USGS operates the satellites and manages data reception, archiving and distribution. Since late 2008 these images have been made available free of charge. The Landsat Program obtains overhead images from a series of satellites. We have created our dataset $\mathcal{D}$ using images from one satellite.  $\mathcal{D}$ consists of 130 color images each cropped at a resolution of $650 \times 650$ pixels. This dataset is further divided into three parts namely training $\mathcal{D}_\text{train}$, validation $\mathcal{D}_\text{val}$ and testing $\mathcal{D}_\text{test}$.

Out of the 130 images, 30 of them are used to create patches for training and validation. Patches of size $64 \times 64 \times 3$ are extracted from every image with a patch stride of $32 \times 32$ generating a total of $10830$ patches. Out of these patches, $20$ percent have been used for validation and the remaining for training. So $\mathcal{D}_\text{train}$ consists of $8664$ patches and $\mathcal{D}_\text{val}$ $2166$ patches.

The remaining 100 images are used for creating $\mathcal{D}_\text{test}$. Half of $\mathcal{D}_\text{test}$ is used to generate forgeries, the remaining half is kept as pristine testing data. In order to create forged images, credible objects such as airplanes, clouds, etc. are spliced at random positions onto the 50 selected images from $\mathcal{D}_\text{test}$. During the splicing operation, the size of spliced objects relative to the used analysis patch size is controlled. Therefore, we define three sizes namely:
\begin{itemize}
	\item Small - Object size is smaller than the patch size (approximately 32 pixel per side).
	\item Medium - Object size is comparable to patch size (approximately 64 pixel per side).
	\item Large - Object size is larger than patch size (approximately 128 pixel per side).
\end{itemize}
Objects of each size are forged onto the 50 images at random positions to create 150 forged images, i.e., we have 50 images with Small objects spliced onto them ($\mathcal{D}_{32}$), 50 images with Medium objects spliced onto them ($\mathcal{D}_{64}$), and at last 50 images with Large objects spliced onto them ($\mathcal{D}_{128}$). Examples of pristine images and forged images with different size forgeries are shown in Figure~\ref{fig:dataset}.

\subsection{Experimental Setup}
Our model consists of two important components, namely the Autoencoder $A$ and the one-class SVM. In this section we describe the policies we used to choose the best model and the various hyper parameters for each of these.

\subsubsection{Autoencoder}
The autoencoder is fed with patches from pristine images and its task is to capture their distribution. An autoencoder can be trained on its own but we choose to couple it with a Discriminator $D$ to further push its training. The job of $D$ is to be able to discriminate between patches produced by $A$ and the actual patches from pristine images. By coupling $A$ with $D$ we form a Generative Adversarial Network (GAN). We can judge the performance of $A$ using the mean squared error metric and the performance of $D$ by binary cross entropy. Both $A$ and $D$ are CNNs and, in order to choose the right CNN architecture, we use the following approach:
\begin{itemize}
	\item The architecture of $D$ is fixed and it is described in the \textit{Feature Extraction} section.
	\item For $A$, a variety of CNN architectures are tested and the one with the lowest mean squared error loss is chosen as the best model.
\end{itemize}

The various architectures tested to choose the best model for $A$ are detailed in Table~\ref{tab:EncArch}


\begin{table*}[t]
\caption{The four proposed autoencoder architectures (shown in columns). The number of filters for each convolutional layer, its size and the used stride is shown in parenthesis followed by the activation function. If no activation is specified for any layer its assumed to be linear activation}
\vspace{0.2cm}
\label{tab:EncArch}
\begin{center}
\begin{tabular}{|c|c|c|c|}
\hline
\multicolumn{4}{|c|}{Encoder and Decoder Architectures}                                                                                                                                                                                                                                 \\ \hline
$A_{1e}$     														 																							& $A_{2e}$                                																												& $A_{3e}$                                  																														& $A_{4e}$                                                \\ \hline
\begin{tabular}{@{}c@{}}conv1 \\ (16, (6,6), Stride 1) + ReLU\end{tabular}           			& \begin{tabular}{@{}c@{}}conv1 \\ (16, (6,6), Stride 1)\end{tabular}  											  & \begin{tabular}{@{}c@{}}conv1 \\ (16, (6,6), Stride 1)\end{tabular}     													& \begin{tabular}{@{}c@{}}conv1 \\ (16, (6,6), Stride 1)\end{tabular}                    \\ \hline
\begin{tabular}{@{}c@{}}conv2 \\ (32, (5,5), Stride 2) + ReLU\end{tabular}           			& \begin{tabular}{@{}c@{}}conv2 \\ (16, (5,5), Stride 2)\end{tabular}  									      & \begin{tabular}{@{}c@{}}conv2 \\ (16, (5,5), Stride 2)\end{tabular}     													& \begin{tabular}{@{}c@{}}conv2 \\ (16, (5,5), Stride 2)\end{tabular}                 \\ \hline
\begin{tabular}{@{}c@{}}conv3 \\ (64, (4,4), Stride 2) + ReLU\end{tabular}           			& \begin{tabular}{@{}c@{}}conv3 \\ (32, (4,4), Stride 2)\end{tabular}                      		& \begin{tabular}{@{}c@{}}conv3 \\ (32, (4,4), Stride 2)\end{tabular}     													& \begin{tabular}{@{}c@{}}conv3 \\ (32, (4,4), Stride 2)\end{tabular}     \\ \hline
\begin{tabular}{@{}c@{}}conv4 \\ (128, (3,3), Stride 2) + ReLU\end{tabular}           			& \begin{tabular}{@{}c@{}}conv4 \\ (32, (3,3), Stride 2)\end{tabular}                        & \begin{tabular}{@{}c@{}}conv4 \\ (32, (3,3), Stride 2)\end{tabular}    													  & \begin{tabular}{@{}c@{}}conv4 \\ (64, (3,3), Stride 2)\end{tabular}     \\ \hline
\begin{tabular}{@{}c@{}}conv5 \\ (256, (2,2), Stride 2) + ReLU\end{tabular}           			& \begin{tabular}{@{}c@{}}conv5 \\ (128, (2,2), Stride 2)\end{tabular}                        & \begin{tabular}{@{}c@{}}conv5 \\ (128, (2,2), Stride 2)\end{tabular}    													& \begin{tabular}{@{}c@{}}conv5 \\ (128, (2,2), Stride 2)\end{tabular}     \\ \hline
\multicolumn{4}{|c|}{BN after each convolutional layer except conv5} \\ \hline
\hline
$A_{1d}$ & $A_{2d}$ & $A_{3d}$ & $A_{4d}$ \\
\hline
\begin{tabular}{@{}c@{}} deconv1 \\ (256, (2,2), Stride 2) + ReLU \end{tabular} & \begin{tabular}{@{}c@{}} deconv1 \\ (32, (2,2), Stride  2) \end{tabular}&\begin{tabular}{@{}c@{}} deconv1 \\ (64, (2,2), Stride  2) \end{tabular}& \begin{tabular}{@{}c@{}} deconv1 \\ (64, (2,2), Stride  2) \end{tabular}\\
\hline
\begin{tabular}{@{}c@{}} deconv2 \\ (128, (3,3), Stride 2) + ReLU \end{tabular} & \begin{tabular}{@{}c@{}} deconv2 \\ (32, (3,3), Stride  2) \end{tabular}&\begin{tabular}{@{}c@{}} deconv2 \\ (32, (3,3), Stride  2) \end{tabular}& \begin{tabular}{@{}c@{}} deconv2 \\ (32, (3,3), Stride 2) \end{tabular}\\
\hline
\begin{tabular}{@{}c@{}} deconv3 \\ (64, (4,4), Stride 2) + ReLU \end{tabular} & \begin{tabular}{@{}c@{}} deconv3 \\ (16, (4,4), Stride  2) \end{tabular}&\begin{tabular}{@{}c@{}} deconv3 \\ (32, (4,4), Stride  2) \end{tabular}& \begin{tabular}{@{}c@{}} deconv3 \\ (16, (4,4), Stride  2) \end{tabular}\\
\hline
\begin{tabular}{@{}c@{}} deconv4 \\ (32, (5,5), Stride 2) + ReLU \end{tabular} & \begin{tabular}{@{}c@{}} deconv4 \\ (16, (5,5), Stride  2) \end{tabular}&\begin{tabular}{@{}c@{}} deconv4 \\ (16, (5,5), Stride  2) \end{tabular}& \begin{tabular}{@{}c@{}} deconv4 \\ (16, (5,5), Stride  2) \end{tabular}\\
\hline
\begin{tabular}{@{}c@{}} deconv5 \\ (3, (6,6), Stride 1) + tanh \end{tabular} & \begin{tabular}{@{}c@{}} deconv5 \\ (3, (6,6), Stride  1) + tanh \end{tabular}&\begin{tabular}{@{}c@{}} deconv5 \\ (3, (6,6), Stride  1) + tanh \end{tabular}& \begin{tabular}{@{}c@{}} deconv5 \\ (3, (6,6), Stride  1) + tanh \end{tabular}\\
\hline
\multicolumn{4}{|c|}{BN after each  deconvolutional layer except deconv5} \\ \hline
\end{tabular}
\end{center}
\end{table*}

 \subsubsection{Training Strategy}
We adopt two different training strategies, \textit{With GAN} and \textit{Without GAN}
 \begin{itemize}
 	\item \textbf{Without GAN}: we train the autoencoder with patches from $\mathcal{D}_\text{train}$ and $\mathcal{D}_\text{val}$. We use the Adam optimizer for a total of 100 epochs. The model weights with the lowest MSE loss on $\mathcal{D}_\text{val}$ over the 100 epochs are chosen as the final model weights.

 	\item \textbf{With GAN}:  we first train the autoencoder using patches from $\mathcal{D}_\text{train}$ and $\mathcal{D}_\text{val}$ using the Adam optimizer for 100 epochs. The weights corresponding to the lowest loss on $\mathcal{D}_\text{val}$ over the 100 epochs are locked. We then use these weights to initialize the weights of the generator in the GAN. The GAN is then trained for 100 epochs with the SGD optimizer and a learning rate of $0.001$ for the discriminator, and the Adam optimizer with a learning rate of $0.001$ for the generator. The GAN training is carried out on batches of 128 patches. The weights with the lowest MSE loss are chosen for the final generator.
\end{itemize}

For the SVM, we used a radial basis function with $\gamma=1 / 2048$ as kernel, and small value of nu-parameter ($\nu = 0.00001$).


\subsection{Results}

Using the \textbf{With GAN} training strategy, the mean squared error (MSE) loss over $\mathcal{D}_\text{val}$ for the various architectures of the autoencoder are reported in Table~\ref{tab:1}. Architectures $A_2$, $A_3$ and $A_4$ have similar number of parameters (about 100k) while $A_1$ has almost a million parameters. Note that $A_1$ , $A_2$ and $A_3$ perform similarly despite the huge difference in the number of parameters. Among all of them $A_4$ provides the lowest MSE and hence better patch reconstruction. Therefore, we decided to select architecture $A_4$ for our system.

\begin{table}[b]
\caption{MSE of the various autoencoder architectures. The one with the lowest MSE loss is selected for the proposed method.} \label{tab:1}
\begin{center}
	\begin{tabular}{ ccc }
		\hline
		\textbf{Architecture} & \textbf{Trainable Parameters} & \textbf{MSE loss} \\
		\hline
		$A_1$ & $997299$ &$0.00131671$ \\ 
		$A_2$ & $84547$ &$0.00131675$  \\  
		$A_3$ & $124883$  &$0.00130047$ \\ 
		$A_4$ & $135939$ &$0.00125511$ \\ 
		\hline
	\end{tabular}
\end{center}
\end{table}

\begin{figure}[t]

	\centering
	\includegraphics[width=.7\columnwidth]{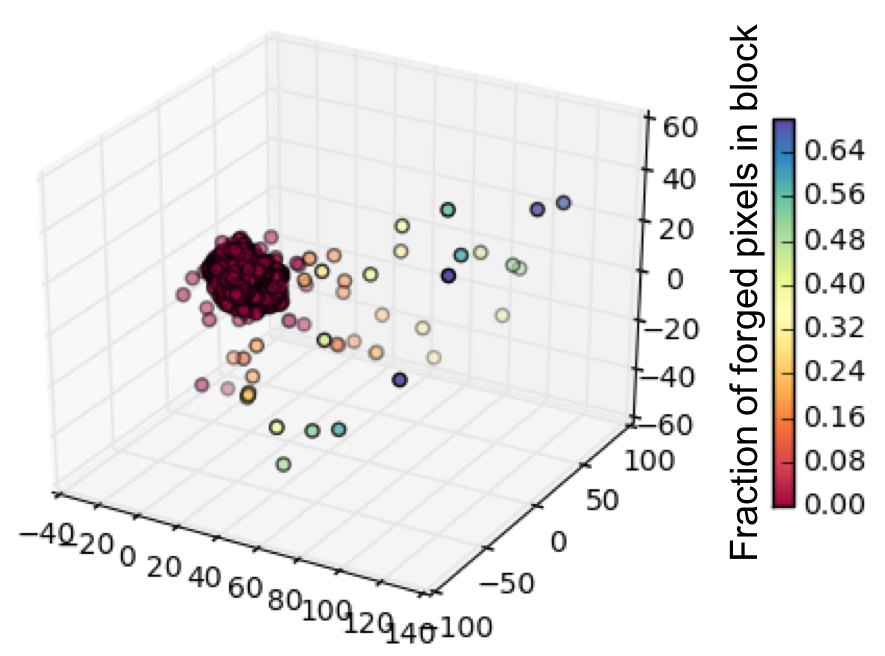}
	\vspace{1em}
	\caption{Example of t-SNE representation of the feature vectors extracted from a forged image. Features from pristine patches (i.e., red dots) cluster together, whereas features from forged patches (i.e., blue dots) are more distant.}
	\label{fig:tsne}
\end{figure}

In order to visualize the forged-vs-pristine discriminability power of feature vectors $\mathbf{h}$ extracted from the proposed autoencoder, we applied the t-SNE algorithm \cite{tsne}. T-SNE can be used for unsupervised feature dimensionality reduction in order to check whether it is possible to cluster some data. In our particular case, the feature vector $\mathbf{h}$ is a high dimensional vector that is very difficult to visualize, whereas, by applying t-SNE we have some visual clues on feature behavior. Figure~\ref{fig:tsne} shows the distribution of features in a reduced dimensionality space of three dimensions using t-SNE. It is possible to notice that patches not containing forged pixels have features that cluster together (i.e., red dots). Conversely, features belonging to patches containing forged pixels (i.e., blue dots) are spread in the three-dimensional space far from the pristine cluster. This confirms that the proposed feature vector is able to capture forged-vs-pristine information.

\begin{table}[b!]
	\centering
	\vspace{2em}
	\caption{Detection results in terms of AUC for the different datasets. AUCs are reported in two different cases: autoencoder trained with or without the GAN. Best results are reported in italics.}
	\vspace{.5em}
	\label{tab:detection}
	\begin{tabular}{cccc}
		\hline
		\textbf{Forgery} & \textbf{AUC}           & \textbf{AUC}        & \textbf{AUC}        \\
		\textbf{Size}    & \textbf{(without Gan)} & \textbf{(with GAN)} & \textbf{Difference} \\ \hline
		Small            & 0.784                 & \textit{0.797}      & +0.013              \\
		Medium           & 0.904                  & \textit{0.920}      & +0.016              \\
		Large            & 0.950                  & \textit{0.972}      & +0.022              \\ \hline
	\end{tabular}
\end{table}

\begin{table}[b!]
	\centering
	\vspace{2em}
	\caption{Localization results in terms of AUC for the different datasets. AUCs are reported in two different cases: autoencoder trained with or without the GAN. Best results are reported in italics.}
	\vspace{.5em}
	\label{tab:localization}
	\begin{tabular}{cccc}
		\hline
		\textbf{Forgery} & \textbf{AUC}           & \textbf{AUC}        & \textbf{AUC}        \\
		\textbf{Size}    & \textbf{(without Gan)} & \textbf{(with GAN)} & \textbf{Difference} \\ \hline
		Small            & \textit{0.913}         & 0.902               & -0.009              \\
		Medium           & \textit{0.963}         & 0.961               & -0.002              \\
		Large            & \textit{0.970}         & 0.974               & -0.004             \\ \hline
	\end{tabular}
\end{table}

\begin{figure}[t]
	\centering
	\subfloat[Forged image $\I$]{\includegraphics[width=.46\columnwidth]{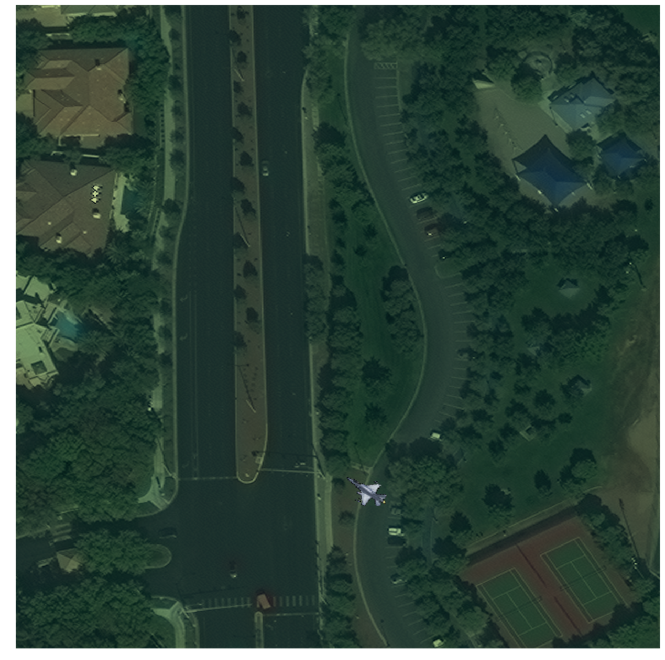}}\hfil
	\subfloat[Forged image $\I$]{\includegraphics[width=.46\columnwidth]{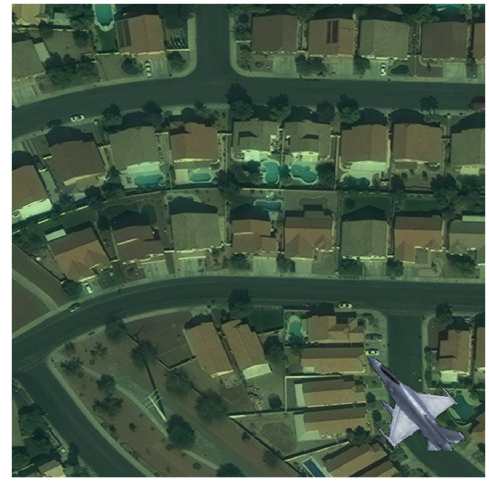}}

	\subfloat[Forged mask $\M$]{\includegraphics[width=.46\columnwidth]{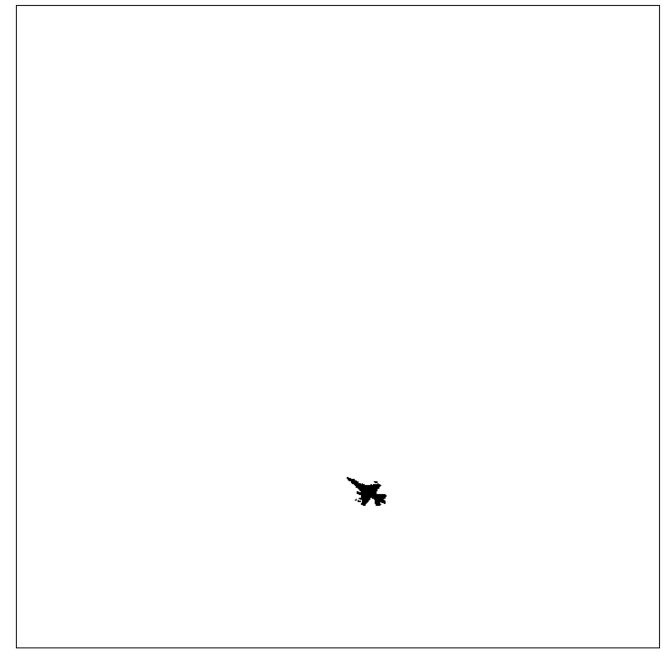}}\hfil
	\subfloat[Forged mask $\M$]{\includegraphics[width=.46\columnwidth]{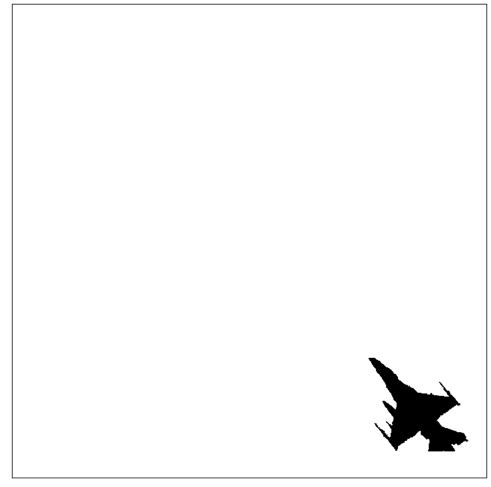}}

	\subfloat[Soft mask $\Mt$]{\includegraphics[width=.46\columnwidth]{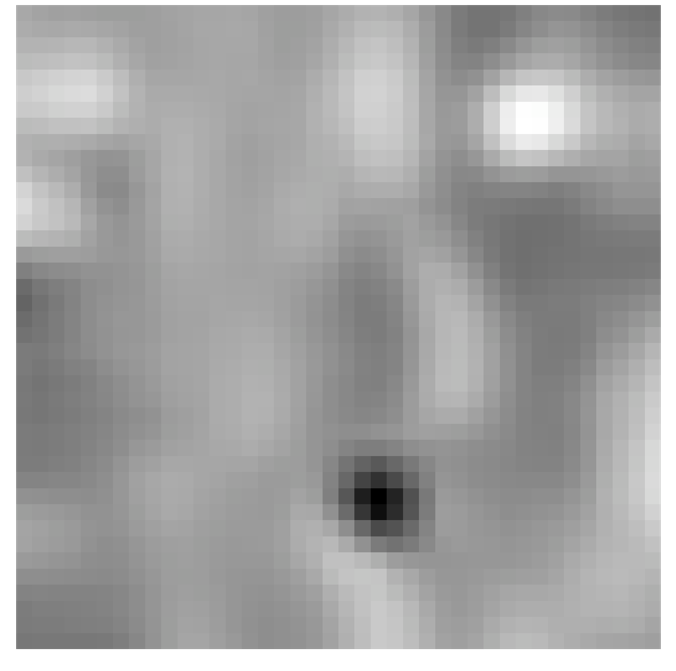}}\hfil
	\subfloat[Soft mask $\Mt$]{\includegraphics[width=.46\columnwidth]{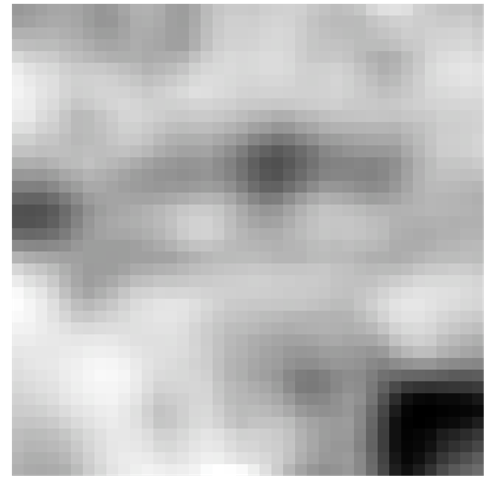}}
	\vspace{1em}
	\caption{Examples of forged images with ground truth forged mask $\M$ and estimated soft mask $\Mt$. It is possible to notice the correlation between ground truth and estimated soft mask.}
	\label{fig:mask}
\end{figure}

\begin{figure}[t]
	\centering
	\includegraphics[width=.85\columnwidth]{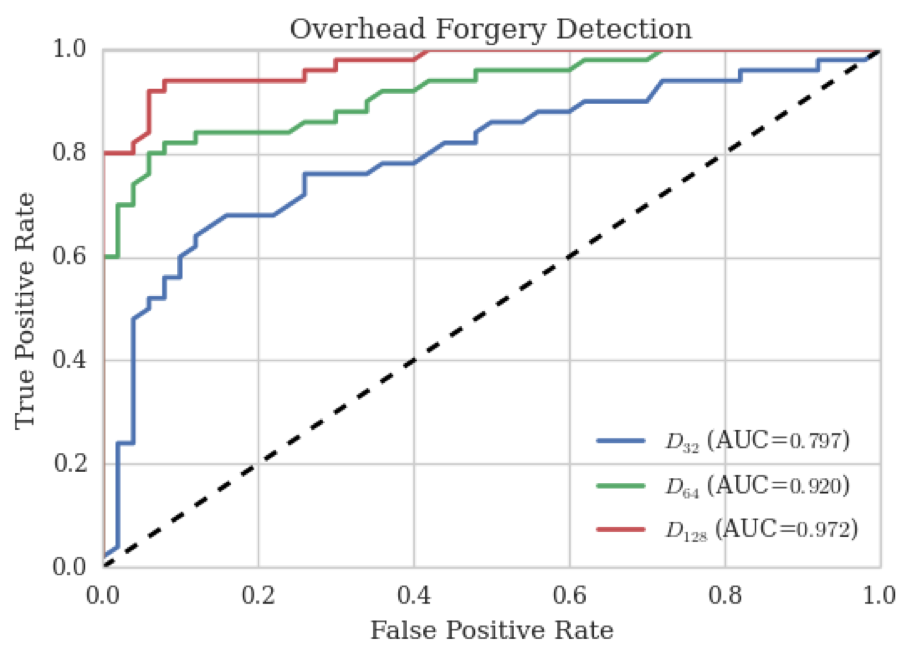}
	\caption{Forgery detection ROC curves. Each curve represents results on a different dataset according to the forgery average size.}
	\label{fig:detection}
\end{figure}

In order to evaluate forgery detection performance, we estimated the soft mask $\Mt$ for each image in the dataset. For each mask, we selected as pristine confidence the minimum $\Mt$ value (i.e., the SVM output associated to the least probable pristine patch). By thresholding this confidence score, we obtained a receiver operating characteristic (ROC) curve. Figure~\ref{fig:mask} shows some examples of forged images, groundtruth masks, and estimated soft mask $\Mt$. Figure~\ref{fig:detection} shows ROC curves split for datasets containing forgeries of different average size. Clearly, the bigger the forgery (i.e., 128 pixel per side), the better the performance (area under the curve around 0.97). However, even when forgeries are smaller than the analysis block (i.e., 32 pixel per side on $64 \times 64$ blocks), the area under the curve (AUC) is almost 0.80.

Additional results are reported in Table~\ref{tab:detection}. This table reports AUC for each different size dataset depending on the used autoencoder training strategy. More precisely, it is possible to notice that by training the autoencoder without the GAN, detection results are always slightly worse. This motivates the use of the GAN training paradigm for forgery detection in this scenario.

\begin{figure}[t]
	\centering
	\includegraphics[width=.85\columnwidth]{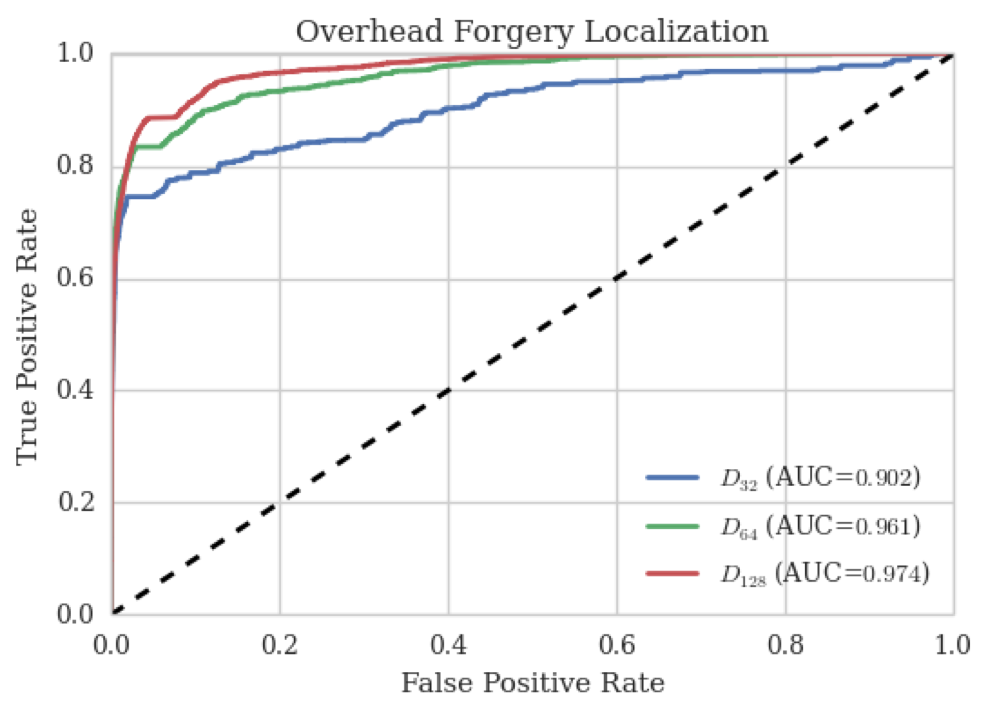}
	\caption{Forgery localization ROC curves. Each curve represents results on a different dataset according to the forgery average size.}
	\label{fig:localization}
\end{figure}

In order to validate the proposed method in terms of localization, we computed a soft mask $\Mt$ for each image in the dataset. We then thresholded each soft mask $\Mt$ to obtain a binary mask soft mask $\Mh$. For each image and used threshold, we computed: the true positive rate as the percentage of forged pixels correctly detected; false positive rate as the percentage of pristine pixels detected as forged. Based on these two values, we drew ROC curves. Figure~\ref{fig:localization} shows ROC curves obtained with our proposed GAN on datasets with forgeries of different size. Specifically, it is possible to notice that AUC is always greater than 0.90. In particular, if the forgeries are twice the size of the analysis patch, the AUC is higher than 0.97.

Additional results are reported in Table~\ref{tab:localization}. We show AUC values for the different datasets (according to forgery sizes), comparing the effect of training the autoencoder with or without the GAN. Notice that, for localization purposes, it is slightly better to avoid the GAN.

A final consideration is devoted to computational time. We tested the proposed algorithm on a workstation equipped with an Intel Core i7-5930K CPU, 128 GB of RAM and a NVIDIA GeForce Titan X GPU. The processing time needed for a $64 \times 64$ pixel patch (considering both the autoencoder and the SVM) was around $500 \mu \mathrm{s}$ for testing. As each patch processing is independent, the algorithm allows for strong parallelization, thus making processing of high resolution images not an issue.

%% file: conclusions.tex
\section{Conclusions}

In this paper we proposed a solution for satellite imagery forgery detection and localization.
The rationale behind the proposed method is that it is possible to train an autoencoder to obtain a compact representation of image patches coming from pristine satellite pictures.
This autoencoder can than be used as a feature extractor for image patches.
During testing, a one-class SVM is used to detect whether feature vectors come from pristine images or not, thus representing forgeries.

The solution proposed in this paper makes use of generative adversarial networks to train the autoencoder for the forgery detection task. 
Moreover, it is worth noting that the whole system is trained only on pristine data.
This means that no prior knowledge on the forgeries is assumed to be available.

Tests on copy-paste attacked images with different forgery size show promising accuracy in both detection and localization.
Future work will be devoted to study system robustness to different kinds of forgeries as well.

%% file: paper.bbl
\begin{thebibliography}{10}

\bibitem{Rocha2011}
A.~Rocha, W.~Scheirer, T.~Boult, and S.~Goldenstein,
\newblock ``Vision of the unseen: Current trends and challenges in digital
  image and video forensics,''
\newblock {\em ACM Computing Surveys}, vol. 43, pp. 1--42, October 2011.

\bibitem{Piva2013}
A.~Piva,
\newblock ``An overview on image forensics,''
\newblock {\em ISRN Signal Processing}, vol. 2013, pp. 22, November 2013.

\bibitem{Stamm2013}
M.~C. Stamm, {Min Wu}, and K.~J.~R. Liu,
\newblock ``Information forensics: An overview of the first decade,''
\newblock {\em IEEE Access}, vol. 1, pp. 167--200, May 2013.

\bibitem{freesat}
GIS Geography,
\newblock {\em 15 Free Satellite Imagery Data Sources}, August 2017 (accessed
  January 1, 2018),
\newblock \emph{http://gisgeography.com/free-satellite-imagery-data-list}.

\bibitem{bbc}
BBC News,
\newblock {\em Conspiracy Files: Who shot down MH17?}, April 2016 (accessed
  January 1, 2018),
\newblock \emph{http://www.bbc.com/news/magazine-35706048}.

\bibitem{russia}
Mashable,
\newblock {\em Satellite images show clearly that Russia faked its MH17
  report}, May 2015 (accessed January 1, 2018),
\newblock \emph{http://mashable.com/2015/05/31/russia-fake-mh17-report}.

\bibitem{Ho2005}
A.~T.~S. Ho, X.~Zhu, and W.~M. Woon,
\newblock ``A semi-fragile pinned sine transform watermarking system for
  content authentication of satellite images,''
\newblock {\em Proceedings of the IEEE International Geoscience and Remote
  Sensing Symposium}, January 2005,
\newblock {Seoul, Korea}.

\bibitem{Ali2017}
L.~Ali, T.~Kasetkasem, F.~G. Khan, T.~Chanwimaluang, and H.~Nakahara,
\newblock ``Identification of inpainted satellite images using evalutionary
  artificial neural network {(EANN)} and k-nearest neighbor {(KNN)}
  algorithm,''
\newblock {\em Proceedings of the IEEE International Conference of Information
  and Communication Technology for Embedded Systems}, May 2017,
\newblock {Chonburi, Thailand}.

\bibitem{Goodfellow2016}
I.~Goodfellow, Y.~Bengio, and A.~Courville,
\newblock {\em Deep Learning},
\newblock MIT Press, Cambridge, MA, 2016.

\bibitem{DBLP:journals/corr/HeZRS15}
K.~He, X.~Zhang, S.~Ren, and J.~Sun,
\newblock ``Deep residual learning for image recognition,''
\newblock {\em arXiv:1512.03385}, December 2015.

\bibitem{DBLP:journals/corr/HeGDG17}
K.~He, G.~Gkioxari, P.~Doll{\'{a}}r, and R.~B. Girshick,
\newblock ``Mask {R-CNN},''
\newblock {\em arXiv:1703.06870v2}, April 2017.

\bibitem{NIPS2012_4824}
A.~Krizhevsky, I.~Sutskever, and G.~E. Hinton,
\newblock ``Imagenet classification with deep convolutional neural networks,''
\newblock {\em Proceedings of the Neural Information Processing Systems
  Conference}, pp. 1097--1105, December 2012,
\newblock Lake Tahoe, NV.

\bibitem{DBLP:journals/corr/RussakovskyDSKSMHKKBBF14}
O.~Russakovsky, J.~Deng, H.~Su, J.~Krause, S.~Satheesh, S.~Ma, Z.~Huang,
  A.~Karpathy, A.~Khosla, M.~S. Bernstein, A.~C. Berg, and L.~Fei-Fei,
\newblock ``Imagenet large scale visual recognition challenge,''
\newblock {\em arXiv:1409.0575v3}, January 2015.

\bibitem{qian2015deep}
Y.~Qian, J.~Dong, W.~Wang, and T.~Tan,
\newblock ``Deep learning for steganalysis via convolutional neural networks,''
\newblock {\em Proceedings of the SPIE/IS\&T Electronic Imaging Conference},
  vol. 9409, pp. 10, January 2015,
\newblock San Francisco, CA.

\bibitem{pibre2015deep}
L.~Pibre, P.~J{\'e}r{\^o}me, D.~Ienco, and M.~Chaumont,
\newblock ``Deep learning for steganalysis is better than a rich model with an
  ensemble classifier, and is natively robust to the cover source-mismatch,''
\newblock {\em Proceedings of the IS\&T International Symposium on Electronic
  Imaging}, vol. 2016, no. 8, February 2016,
\newblock San Francisco, CA.

\bibitem{Xu2016}
G.~Xu, H.~Z. Wu, and Y.~Q. Shi,
\newblock ``Structural design of convolutional neural networks for
  steganalysis,''
\newblock {\em IEEE Signal Processing Letters}, vol. 23, no. 5, pp. 708--712,
  March 2016.

\bibitem{Sedighi2017}
V.~Sedighi and J.~Fridrich,
\newblock ``Histogram layer, moving convolutional neural networks towards
  feature-based steganalysis,''
\newblock {\em Proceedings of the IS\&T International Symposium on Electronic
  Imaging}, vol. 2017, no. 7, January 2017,
\newblock Burlingame, CA.

\bibitem{Chen2015}
C.~Jiansheng, K.~Xiangui, L.~Ye, and Z.~J. Wang,
\newblock ``Median filtering forensics based on convolutional neural
  networks,''
\newblock {\em IEEE Signal Processing Letters}, vol. 22, no. 11, pp.
  1849--1853, June 2015.

\bibitem{Bayar2016}
B.~Bayar and M.~C. Stamm,
\newblock ``A deep learning approach to universal image manipulation detection
  using a new convolutional layer,''
\newblock {\em Proceedings of the ACM Workshop on Information Hiding and
  Multimedia Security}, pp. 5--10, June 2016,
\newblock {Vigo, Spain}.

\bibitem{Tuama2016a}
A.~Tuama, F.~Comby, and M.~Chaumont,
\newblock ``Camera model identification with the use of deep convolutional
  neural networks,''
\newblock pp. 1--6, December 2016,
\newblock {Abu Dhabi, United Arab Emirates}.

\bibitem{Bondi2017}
L.~Bondi, L.~Baroffio, D.~G\"{u}era, P.~Bestagini, E.~J. Delp, and S.~Tubaro,
\newblock ``First steps toward camera model identification with convolutional
  neural networks,''
\newblock {\em IEEE Signal Processing Letters}, vol. 24, no. 3, pp. 259--263,
  March 2017.

\bibitem{Bondi2017a}
L.~Bondi, S.~Lameri, D.~G\"{u}era, P.~Bestagini, E.~J. Delp, and S.~Tubaro,
\newblock ``Tampering detection and localization through clustering of
  camera-based {CNN} features,''
\newblock {\em Proceedings of the IEEE Conference on Computer Vision and
  Pattern Recognition Workshops}, pp. 1855--1864, July 2017,
\newblock {Honolulu, HI}.

\bibitem{Guera2017}
D.~G\"{u}era, Y.~Wang, L.~Bondi, P.~Bestagini, S.~Tubaro, and E.~J. Delp,
\newblock ``A counter-forensic method for {CNN}-based camera model
  identification,''
\newblock {\em Proceedings of the IEEE Conference on Computer Vision and
  Pattern Recognition Workshops}, pp. 1840--1847, July 2017,
\newblock {Honolulu, HI}.

\bibitem{wang2016double}
Q.~Wang and R.~Zhang,
\newblock ``Double {JPEG} compression forensics based on a convolutional neural
  network,''
\newblock {\em EURASIP Journal on Information Security}, vol. 2016, no. 1, pp.
  23, December 2016.

\bibitem{Barni2017}
M.~Barni, L.~Bondi, N.~Bonettini, P.~Bestagini, A.~Costanzo, M.~Maggini,
  B.~Tondi, and S.~Tubaro,
\newblock ``Aligned and non-aligned double {JPEG} detection using convolutional
  neural networks,''
\newblock {\em Journal of Visual Communication and Image Representation}, vol.
  49, no. Supplement C, pp. 153--163, November 2017.

\bibitem{BN}
S.Ioffe and C.~Szegedy,
\newblock ``Batch normalization: Accelerating deep network training by reducing
  internal covariate shift,''
\newblock {\em arXiv:1502.03167v3}, March 2015.

\bibitem{Cozzolino2016}
D.~Cozzolino and L.~Verdoliva,
\newblock ``Single-image splicing localization through autoencoder-based
  anomaly detection,''
\newblock {\em Proceedings of the IEEE International Workshop on Information
  Forensics and Security}, pp. 1--6, December 2016,
\newblock Abu Dhabi, United Arab Emirates.

\bibitem{DAvino2017}
D.~D'Avino, D.~Cozzolino, G.~Poggi, and L.~Verdoliva,
\newblock ``Autoencoder with recurrent neural networks for video forgery
  detection,''
\newblock {\em Proceedings of the IS\&T Electronic Imaging Conference}, vol.
  2017, no. 7, January 2017,
\newblock Burlingame, CA.

\bibitem{Pathak2016}
D.~Pathak, P.~Krahenbuhl, J.~Donahue, T.~Darrell, and A.~Efros,
\newblock ``Context encoders: Feature learning by inpainting,''
\newblock {\em Proceedings of the IEEE Conference on Computer Vision and
  Pattern Recognition}, pp. 2536--2544, June 2016,
\newblock Las Vegas, NV.

\bibitem{NIPS2014_5423}
I.~Goodfellow, J.~Pouget-Abadie, M.~Mirza, B.~Xu, D.~Warde-Farley, S.~Ozair,
  A.~Courville, and Y.~Bengio,
\newblock ``Generative adversarial nets,''
\newblock {\em Proceedings of the Neural Information Processing Systems
  Conference}, pp. 2672--2680, December 2014,
\newblock Montr\'{e}al, Canada.

\bibitem{landsat}
Amazon Web~Services Inc.,
\newblock {\em Landsat on AWS}, Accessed January 1, 2018,
\newblock \emph{https://aws.amazon.com/public-datasets/landsat/}.

\bibitem{landsat2}
National Aeronautics and Space Administration,
\newblock {\em Landsat Science}, Accessed January 1, 2018,
\newblock \emph{https://landsat.gsfc.nasa.gov/}.

\bibitem{NASA}
National Aeronautics and Space Administration,
\newblock {\em NASA}, Accessed January 1, 2018,
\newblock \emph{https://www.nasa.gov/}.

\bibitem{USGS}
U.S.~Geological Survey,
\newblock {\em USGS.gov | Science for a changing world}, Accessed January 1,
  2018,
\newblock \emph{https://www.usgs.gov/}.

\bibitem{tsne}
L.~van~der Maaten and G.~Hinton,
\newblock ``Visualizing high-dimensional data using t-{SNE},''
\newblock {\em Journal of Machine Learning Research}, vol. 9, pp. 2579--2605,
  November 2008.

\end{thebibliography}
